\documentclass{article}
\usepackage{amsmath}
\usepackage{mathtools}
\usepackage{graphicx}
\usepackage{float}
\begin{document}

\title{A Supervised Modified Hebbian Learning Method On Feed-forward Neural Networks}
\author{Rafi Qumsieh \\  rqumsieh@columbustech.edu}

\maketitle

\begin{abstract}
In this paper, we present a new supervised learning algorithm that is based on the Hebbian learning algorithm in an attempt to offer a substitute for back propagation along with the gradient descent for a more biologically plausible method. The best performance for the algorithm was achieved when it was run on a feed-forward neural network with the MNIST handwritten digits data set reaching an accuracy of  70.4\% on the test data set and 71.48\% on the validation data set.
\end{abstract}

\section{Introduction and Motivation}

Back propagation along with gradient descent are the primary algorithms used in training state-of-the-art neural network models. They have been very successful in producing very efficient models. They work by attempting to minimize the cost function by rolling down the cost function using gradient descent which depends on calculating partial derivatives with respect to the weights and biases through what is known as the back propagation process. A couple of the concerns regarding these algorithms are:

\begin{enumerate}
	\item They are not biologically plausible as we have no evidence of back propagation happening in neural systems.
	\item They are computationally expensive as they need to calculate a potentially large number of partial derivatives with respect to weights and biases for each data point.
\end{enumerate}

Even Geoffrey Hinton, one of the founders of modern Artificial Intelligence remarked that he is suspicious of back propagation and that we ought to start over. [1]

Donald Hebb, a Canadian psychologist, postulated that the brain is plastic and it learns through changing the synaptic connections strengths between neurons depending on whether the input signal caused the output neuron to fire or not, and how large the signals are [2]. If a neuron causes another neuron to fire, their connection is strengthened in what is known as the long-term potentiation process. If a neuron fires, but does not cause the other neuron to fire, the connection weakens in what is known as the long-term depression process. Mathematically, we can describe Hebb's postulate as the following: 
\begin{equation}
	\Delta w = \eta x y
\end{equation}
Where $w$ is the connection weight: a numerical value that indicates the strength of the connection between the two neurons, $\eta$ is the learning rate: a small positive number that indicates how much the weight will change at each iteration, $x$ is the input signal, $y$ is the output of the neuron. This formula captures some parts of Hebb's hypothesis as it increases the weight if both $x$ and $y$ are large. It does not change the weight if any or both of them are zero. It might have a couple of problems:

\begin{enumerate}
	\item If the input $x$ and output $y$ are large, the weight change can grow indefinitely. This is something that is not plausible in nature.
	\item If one of them is zero while the other is not, the formula returns zero change for the weight change, but evidence from the long-term depression process shows that if the input is large but the neuron does not cause an output signal in the following neuron, a metabolic process that reduces the synaptic weight will take place.
\end{enumerate}

A solution to the first problem was introduced by Oja in what is known as Oja's rule [3]. It is a modified version of Hebb's rule where the weight vector to each neuron does not change its magnitude, only its direction. One solution to the second problem is to treat -1 values as zeroes. But instead of relying on one formula to capture a set of potentially complex processes in nature, namely long-term potentiation and long-term depression, we will write an algorithm that handles the different scenarios that the neurons face while learning and how they react to them.\\

In this paper, we discuss a feed-forward neural network that is trained on the MNIST handwritten digits data set using a modified Hebbian learning algorithm.

\section{The Modified Hebbian Learning Algorithm}

The algorithm is used to train a neural network using the  training data set by taking each training data point $(x_{train},y_{train})$ and running the following:

\begin{enumerate}
	
	\item Feed the signal forward by calculating the activation of each neuron in the network using a rectified version of the hyperbolic tangent activation function $ y = \tanh_{rec}(\Sigma x_i w_i - b_i) $. This $\tanh_{rec}$ function takes a hyper parameter $c_{activation} $ that is the coefficient of $x$ and it controls the curve of $\tanh$. At the every layer, the activation of the layer is appended to an activations list. At the last layer, the desired output vector is forcefully appended to the activations list (in a winner-takes-all manner) so that we ensure a supervised learning of the pattern. The activation function of the last output layer is the $ReLU$ function to make distinguishing values easier (instead of squashing the values). The $\tanh_{rec}$ function has the following definition:
	
		\begin{equation}
	\tanh_{rec}(x, c) = \left\{\begin{array}{lr}
	\frac{e^{cx}-e^{-cx}}{e^{cx}+e^{-cx}}, & \text{if } x > 0  \\ \\
	0 & \text{if } x \leq 0 \\ \\
	\end{array}\right\}  
	\end{equation}
	
	\item After obtaining all activations for the data point, for each weight, we evaluate the modified Hebbian weight update function asynchronously (with respect to each layer), starting from the first layer:

	\begin{equation}
	\Delta w(x,y) = \left\{\begin{array}{lr}
	+\eta_{ltp} x y, & \text{if } x * y \geq T, w \neq 0  \\ \\
	-\eta_{ltp} x y, & \text{if } x * y < T, w \neq 0  \\ \\
	0.50, & \text{if } w = 0  \\ \\
	\end{array}\right\}
	\end{equation}	

Where $T$ is a positive number that represents the threshold at which the weight will change for each neuron. $x$ and $y$ have values between 0 and 1. Sometimes, we modify this weight update function to handle additional cases. The above weight update function is a compressed version that captures enough information for the neuron to learn using the long-term potentiation and long-term depression processes. We can expand the weight update function to include more cases as we try model more processes / scenarios from biology. Below is an extended version of the function:
	\begin{equation}
	\Delta w(x,y) = \left\{\begin{array}{lr}
	+\eta_{ltp} x y, & \text{if } x > 0 ,\space y > 0,\space w > 0  \\ \\
	-\eta_{ltp} x y, & \text{if } x > 0 ,\space y > 0, \space w < 0  \\ \\
	-\eta_{ltd} x, & \text{if } x>0, \space y=0,\space w > 0 \\ \\
	+\eta_{ltd} x, & \text{if } x>0, \space y=0,\space w < 0 \\ \\
	+\eta_{ltp2} y, & \text{if } x=0, \space y>0,\space w > 0 \\ \\
	-\eta_{ltp2} y, & \text{if } x=0, \space y>0,\space w < 0 \\ \\
	+0.50 & \text{if } x*y\geq T,\space w=0 \\ \\
	0, & \text{Otherwise} \\ \\
	\end{array}\right\}
	\end{equation}	
	
	\item Calculate the new updated weights using the rectified linear unit function $ReLU(x)$ as follows: $w_{new} = ReLU(w_{old} + \Delta w) $. If $w_{old} > 0$ but $w_{new} < 0 $, then reset $w$ to zero and vice versa. The rationale here is that a neuron cannot change its type from being excitatory to being inhibitory, or the other way around. The rectified linear unit function is defined as:
	\begin{equation}
	  	ReLU(x) = \left\{\begin{array}{lr}
	 x, & \text{if } x > 0  \\ \\
	  0 & \text{if } x \leq 0 \\ \\
	  \end{array}\right\}  
	\end{equation}

\end{enumerate}

	Note that these rules come from assumptions and attempts to model some well known mechanisms on how plasticity works. We can always modify these rules as we discover more information on how plasticity, long-term potentiation, and long-term depression work.

\section{The Networks}

\subsection{The Shallow Network}

\subsubsection{Structure Of The Network}

First, let us start with describing the problem. We are trying to classify a grayscale image of size 28 by 28 pixels of a handwritten digit to one of the digits between 0 and 9. One way to look at the problem is that we are trying to find a classifier map $C$ that can be defined as follows:
\begin{equation}
C : [0,1]^{784} \to \{0,1\}^{10}
\end{equation}

We will represent the desired map by creating a feedforward neural network with two layers: An input layer of size 784 and an output layer of size 10 and no hidden layers. In this paper, a feed-forward neural network  written in Python by Michael Nielsen [4] was used with the following layers:

\begin{enumerate}
	\item Input layer: This layer has 784 neurons corresponding to the cells in the input of the 28 by 28 pixel gray-scale images used in the MNIST data set. Each neuron's output is a real number between 0 and 1.
	\item Output layer: This layer has 10 neurons. Each neuron corresponds to a digit from 0 to 9. The neuron that outputs the highest value is interpreted as being the classified neuron.
\end{enumerate}

\subsubsection{Weights And Biases Initialization}

All neurons in the network had a fixed bias of 0.95 that was adjusted along with the other hyper parameters. The idea behind having a fixed bias is that, as far as we know, neurons have a fixed threshold voltage of ~ -70 mV, so our assumption brings us closer to nature. As for the weights, the connection weights have the following configuration:

\begin{enumerate}
	\item Input-to-Output layer: All weights are initialized to a preset value. In our case, we can start the weight with value $ w = 0$. Having all weights equal zero gives the network no prejudice towards any output as it starts learning.
\end{enumerate}

\subsubsection{Properties Of All The Variables And Parameters}

\begin{enumerate}
	\item $x$ is the input signal for each neuron. $x \in [0,1]$. Where a value of 0 indicates no signal, and a value of 1 indicates maximum signal.
	\item $y$ is the output signal for each neuron. $y \in [0,1]$. Where a value of 0 indicates no signal, and a value of 1 indicates maximum output signal.
	\item $w$ is the connection weight value for each pair of neurons from a layer to a consecutive layer. $w \in [-1,1]$. Where a negative value indicates that the connection is inhibitory and a positive value indicates an excitatory connection. A value closer to zero indicates a weaker connection. Anytime an adjusted weight exceeds the value of 1, it will be either passed through a squashing function or hard-resetted to the value of 0.90, and similarly, anytime an adjusted weight value decreases beyond -1, it will be either passed through a squashing function or hard-resetted to -0.90. This technique is used to ensure that not all weights reach complete saturation, and thus will hinder learning. Even though the weights can have negative values, we restricted the values to positive ones in the experiments.
\end{enumerate}

\subsection{The Medium Network}

The previous shallow network can do a pretty decent job at mapping input vectors to output vectors. It faces the problem of needing to learn a relatively large number of connections between the input and the output layers (the input is large), and this will cause a long learning time as we will see in the results section. We can add a pooling / dimensionality reduction layer that extracts important features from the input, and then learn the new reduced input. This is the purpose of using the medium network.
	
\subsubsection{Structure Of The Network}

The network has the following layers:

\begin{enumerate}
	\item Input layer: This layer has 784 neurons corresponding to the cells in the input of the 28 by 28 pixel gray-scale images used in the MNIST data set.
	\item Hidden layer: This layer can have either 196 or 49 neurons depending on how much we want to reduce the input by.
	\item Output layer: This layer has 10 neurons. Each neuron corresponds to a digit from 0 to 9.
\end{enumerate}

\subsubsection{Weights And Biases Initialization}

All neurons in the same layer had a fixed bias that was adjusted along with the other hyper parameters. As for the weights, the connection weights have the following configuration:

\begin{enumerate}
	\item Input-to-Hidden layer: The weights are initialized to values $ w \in [-1, 1]$ with any desired probability distribution, and the connections are set up in a configuration that forces dimensionality reduction on the input data through a form of pooling. More on this in the Appendix A. In the experiments, this layer was not set to learn and so its weights did not change. 
	
	\item Hidden-to-Output layer: The weights can start with any fixed value in $ w \in [-1, 1]$  that can be changed along with other hyper parameters, and the connections are set up in a way that resembles a fully connected configuration.
\end{enumerate}

One way to view the structure of this network is that the weights are set up in a manner that resembles a pooling map that reduces dimensionality, followed by a map of association between the reduced input and the desired output.

\subsubsection{Properties Of All The Variables And Parameters}

\begin{enumerate}
	\item $x$ is the input signal for each neuron. $x \in [0,1]$. Where a value of 0 indicates no signal, and a value of 1 indicates maximum signal.
	\item $y$ is the output signal for each neuron. $y \in [0,1]$. Where a value of 0 indicates no signal, and a value of 1 indicates maximum output signal.
	\item $w$ is the connection weight value for each pair of neurons from a layer to a consecutive layer. $w \in [-1,1]$. Where a negative value indicates that the connection is inhibitory and a positive value indicates an excitatory connection. A value closer to zero indicates a weaker connection. Anytime an adjusted weight exceeds the value of 1, it will be either passed through a squashing function or hard-resetted to the value of 0.90, and similarly, anytime an adjusted weight value decreases beyond -1, it will be either passed through a squashing function or hard-resetted to -0.90. This technique is used to ensure that not all weights reach complete saturation, and thus will hinder learning.
\end{enumerate}

\subsection{The Deeper Network}

The previous medium (3-layered) network reduces the dimensionality of the input which theoretically should speed up the learning process. We wanted to see if we can reduce the learning time even further while allowing the network to learn more. This is the purpose of this network.
	
\subsubsection{Structure Of The Network}

The network has the following layers:

\begin{enumerate}
	\item Input layer: This layer has 784 neurons corresponding to the cells in the input of the 28 by 28 pixel gray-scale images used in the MNIST data set.
	\item Hidden layer 1: This layer can have either 196 or 49 neurons depending on how much we want to reduce the input by.
	\item Hidden layer 2: This layer can have either 196 or 49 neurons depending on how much we want to reduce the input by. We prefer this layer to have a smaller number of neurons than hidden layer 1.
	\item Output layer: This layer has 10 neurons. Each neuron corresponds to a digit from 0 to 9.
\end{enumerate}

\subsubsection{Weights And Biases Initialization}

All neurons in each layer have a fixed bias value that can be adjusted along with other hyper parameters. As for the weights, the connection weights have the following configuration:

\begin{enumerate}
	\item Input-to-Hidden 1 layer : The weights are initialized to values $ w \in [-1, 1]$ with any desired probability distribution, and the connections are set up in a configuration that forces dimensionality reduction on the input data through a form of pooling. More on this in the Appendix A. In the experiments, this layer was not set to learn and so its weights did not change. 

	\item Hidden 1-to-Hidden 2 layer : The weights are initialized to values $ w \in [-1, 1]$ with any desired probability distribution, and the connections are set up in a configuration that forces dimensionality reduction on the input data through a form of pooling. More on this in the Appendix A. In the experiments, this layer was not set to learn and so its weights did not change. 
	
	\item Hidden 2-to-Output layer: The weights can start with any fixed value in $ w \in [-1, 1]$  that can be adjusted along with other hyper parameters, and the connections are set up in a way that resembles a fully connected configuration.
\end{enumerate}

One way to view the structure of this network is that the weights are set up in a manner that resembles a pooling map that reduces dimensionality, followed by another pooling map that reduces dimensionality and extracts important features, followed by a map of association between the final reduced input and the desired output.

\subsubsection{Properties Of All The Variables And Parameters}

\begin{enumerate}
	\item $x$ is the input signal for each neuron. $x \in [0,1]$. Where a value of 0 indicates no signal, and a value of 1 indicates maximum signal.
	\item $y$ is the output signal for each neuron. $y \in [0,1]$. Where a value of 0 indicates no signal, and a value of 1 indicates maximum output signal.
	\item $w$ is the connection weight value for each pair of neurons from a layer to a consecutive layer. $w \in [-1,1]$. Where a negative value indicates that the connection is inhibitory and a positive value indicates an excitatory connection. A value closer to zero indicates a weaker connection. Anytime an adjusted weight exceeds the value of 1, it will be either passed through a squashing function or hard-resetted to the value of 0.90, and similarly, anytime an adjusted weight value decreases beyond -1, it will be either passed through a squashing function or hard-resetted to -0.90. This technique is used to ensure that not all weights reach complete saturation, and thus will hinder learning.
\end{enumerate}

\section{Setup}	

The experiments were performed on an HP\textsuperscript{\textregistered} Pavilion laptop with an i5 processor and 8 GB RAM memory on Windows\textsuperscript{\textregistered}. Everything was written using Python 3.7 on top of Anaconda. The code was imported and modified from Michael Nielsen's neural network book's code [2]. All of the code involved in this experiment can be found using reference [5].

\section{The Experiments And Results}	

There were two objectives in the experiments. The first objective was to achieve high accuracies using a very small number of images per digit (Low-shot learning), the second objective was to achieve the highest possible accuracy. The experiments involved tuning many parameters, but below are the optimal hyper parameters found for each type of network used:

\subsection{The Shallow Network}

\subsubsection{Low-Shot Learning}

Using only 2 images per digit over 1 epoch, the network scored an accuracy of 47.62\% on the test data.

\subsubsection{Highest Accuracy}

For this part, the optimal hyper parameters that resulted in the highest prediction accuracy on the test and validation data are:

\begin{enumerate}
	\item The feed-forward neural network had 3 layers of sizes 784, 196, 10 respectively.
	\item $ \eta_{ltp} = 0.001 $ and $ \eta_{ltd} = 0.0001 $.
	\item $ c_{output} = 0.05 $ and $ c_{weights} = 0.5 $.
	\item Biases of neurons in each layer = [0.95]
	\item Initial weights of neurons in each layer = [0.00]
	\item Weight creation value is 0.50 at threshold $x*y \geq 0.25$
	\item Number of images per digit = 60 images per digit which yields 600 unique images.
	\item Number of epochs = 1
\end{enumerate}

Using this configuration, the network gave about 81\%, 69\%, 71.05\% on the training, test, validation datasets respectively.

\subsubsection{Overall trend}
The network needed a training time of about 60.76 seconds for the 600 training data points, and its accuracy vs. IPD relationship is graphed below:

\begin{figure}[H]
  \includegraphics[width=\linewidth]{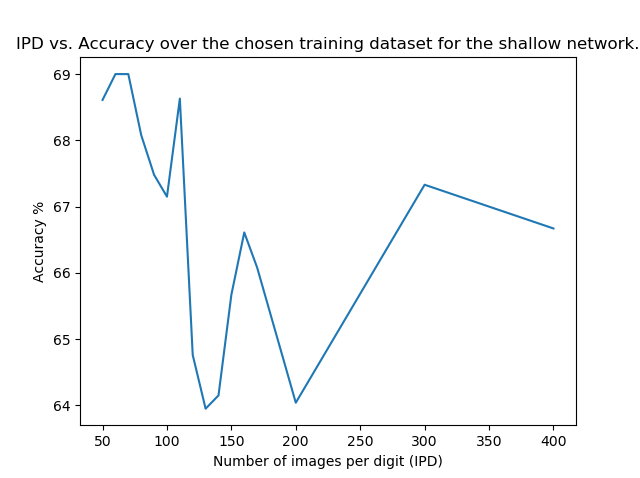}
  \caption{The accuracy of the shallow network as a function of the number of images-per-digit (IPD).}
  \label{fig:shallownetwork}
\end{figure}

It can be seen that there is an overall decreasing trend for the accuracy as we increase the IPD value. 

\subsection{The Medium Network}

\subsubsection{Low-Shot Learning}

Using only 1 image per digit over 1 epoch, the network scored an accuracy of 36.25\% on the test data, and using 2 images per digit over 1 epoch, the network gave an accuracy of 43.63\% on the test data.

\subsubsection{Highest Accuracy}

For this part, the optimal hyper parameters that resulted in the highest prediction accuracy on the test and validation data are:

\begin{enumerate}
	\item The feed-forward neural network had 3 layers of sizes 784, 196, 10 respectively.
	\item $ \eta_{ltp} = 0.01 $ and $ \eta_{ltd} = 0.0005 $.
	\item $ c_{output} = 0.50 $ and $ c_{weights} = 0.50 $.
	\item biases of neurons in each layer = [0.95, 0.00]
	\item Initial weights of neurons in each layer = [0.50, 0.00]
	\item Weight creation value is 0.50 at threshold $x*y \geq 0.25$
	\item Number of images per digit = 60 images per digit which yields 600 unique images.
	\item Number of epochs = 1
\end{enumerate}

Using this configuration, the network gave about 79.0\%, 70.4\%, 71.48\% on the training, test, validation datasets respectively.

\subsubsection{Overall Trend}

The network needed a training time of about 14.18 seconds to train on 600 data points, and its accuracy vs. IPD relationship is graphed below:

\begin{figure}[H]
  \includegraphics[width=\linewidth]{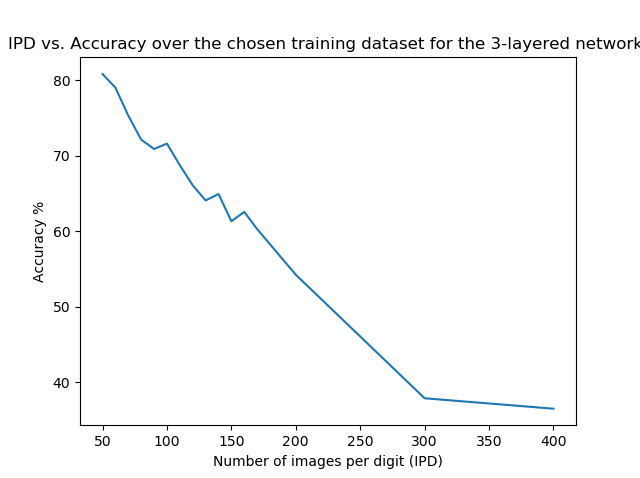}
  \caption{The accuracy of the medium network as a function of the number of images-per-digit (IPD).}
  \label{fig:med}
\end{figure}

\subsection{The Deeper Network}

\subsubsection{Low-Shot Learning}

Using only 1 image per digit over 1 epoch, the network scored an accuracy of 36.33\% on the test data.

\subsubsection{Highest Accuracy}

For this part, the optimal hyper parameters that resulted in the highest prediction accuracy on the test and validation data are:

\begin{enumerate}
	\item The feed-forward neural network had 3 layers of sizes 784, 196, 10 respectively.
	\item $ \eta_{ltp} = 0.001 $ and $ \eta_{ltd} = 0.0001$.
	\item $ c_{output} = 1.0 $ and $ c_{weights} = 0.5 $.
	\item biases of neurons in each layer = [0.35, 0.05, 0.00]
	\item Initial weights of neurons in each layer = [0.50, 0.60, 0.00]
	\item Weight creation value is 0.50 at threshold $x*y \geq 0.25$
	\item Connectivity factors in each layer = [0.75, 0.25]
	\item Number of images per digit = 200 images per digit which yields 2,000 unique images.
	\item Number of epochs = 1
\end{enumerate}

Using this configuration, the network gave about 58.33\%, 48.17\%, 50.27\% on the training, test, validation datasets respectively.

\subsubsection{Overall Trend}

The network needed a training time of about 7.72 seconds to train on 600 data points, and its accuracy vs. IPD relationship is graphed below:

\begin{figure}[H]
  \includegraphics[width=\linewidth]{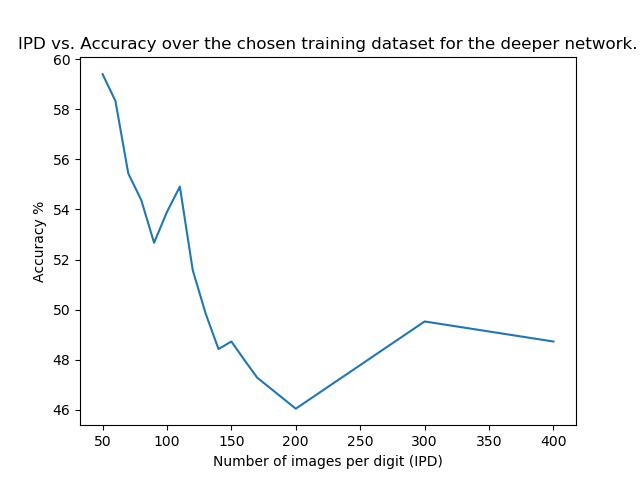}
  \caption{The accuracy of the deeper network as a function of the number of images-per-digit (IPD).}
  \label{fig:deeper}
\end{figure}

\section{Discussion and Conclusion}	

Testing the learning algorithm across the different types of networks and across the different test sets shows that there is some form of learning taking place in those networks. The networks seem to reach a saturation point where adding more training data does not increase the accuracy, but decrease it instead. The highest achieved accuracy is about 70\% for the test data on the medium (3-layered) network. Increasing the number of epochs/recitations seems to increase the accuraries only for the smallest number of IPD (1 - 10), but seems to decrease the accuracies for larger numbers of IPD. One possible reason for the saturation is that none of the networks had any inhibitory neurons (negative weights) which can help regulate the network and prevent it from reaching saturation with a small number of training data points. Another thing that is worth mentioning is that using the detailed version of the algorithm versus the compressed version did not seem to have a large impact on the accuracy measurements. \\

One way to increase the accuracies even further is to use what can be referred to as "assisted learning". Assisted learning basically involves looking at which labels were poorly classified by the algorithm, and offering more data points for the labels that were most poorly classified. So for example, if our model did not classify the digit '2' well, we would add more data points for the label '2' during training. This is similar to helping a student who is struggling with a certain concept by having them work on more exercises that involve that topic until they master it. Keep in mind that the method of assisted learning is not specific to this algorithm and can probably be applied to different models. \\

A potential improvement to the network might lie in adding another layer that takes the reduced input and maps it to a higher dimension in a way that preserves its main features but makes differentiation between inputs easier for the final fully connected layer.\\

The algorithm discussed in this paper has the following advantages:

\begin{enumerate}
	\item This learning algorithm attempts to mimic our current understanding of the processes that underlie plasticity. This is especially beneficial for neuromorphic chips that use memristors which work in a similar manner.
	\item The algorithm does not require the model to have random values for weights and biases, which is something that is more likely not found in nature as cells are created using specific instructions.
	\item The learning occurs simultaneously with the feedforward process in a less computationally expensive way.
\item It can be highly modified to handle more cases and scenarios as we learn more about plasticity.
\end{enumerate}

On the other hand, the algorithm at this point has the following disadvantages and caveats:

\begin{enumerate}
	\item It needs to be more formally defined.
	\item It needs thorough testing and evaluating.
	\item Inhibitory neurons / negative weights should be incorporated into the model as a way control information flow.
	\item Learning only occurs at the last layer where we force the output to be the desired output. If learning can be extended to more layers, it will probably give the network a larger capacity to store more information.
\end{enumerate}

There are still many questions and issues that need to be addressed to further the understanding of the algorithm. This paper serves as a pilot to pique interest in the algorithm and its potential applications.

\section{References}
\begin{enumerate}
\item Jeoffrey Hinton's interview with Axios - Axios - https://www.axios.com/artificial-intelligence-pioneer-says-we-need-to-start-over-1513305524-f619efbd-9db0-4947-a9b2-7a4c310a28fe.html
\item The Organization of Behavior - Donald Hebb - Wiley \& Sons.
\item Simplified neuron model as a principal component analyzer - Erkki Oja - Journal of Mathematical Biology. 15 (3): 267–273
\item Neural Networks and Deep Learning - Michael Nielsen - http://neuralnetworksanddeeplearning.com
\item Algorithm's Code Repository - Rafi Qumsieh - https://github.com/rqumsieh0/supervised-hebbian
\end{enumerate}
\newpage
\appendix

\section{Appendix A - Details Of The Pooling Map}

Let us assume we have a layer (set) of cells of size $n$, and let us call this set $C^n$. Now, let us define a new layer (set) of cells of size $m$, and let us call it $C^m$. Associated with each cell in these sets a value $b$ that denotes its bias to firing given an input. Let the following assumptions be true about our system:

\begin{enumerate}
	\item $n$ is a perfect square, and $m$ is a perfect square. Furthermore, $m = \sqrt{n}$
	\item The first layer is the input layer, and that information flows uni-directionally from the first layer to the second layer only.
	\item The cells in each layer are not connected to each other.
\end{enumerate}

It is important to notice that any of the layers can be arranged into a square matrix. For example, $C^n$ can be arranged into a square matrix of size $m \times m$, and the row and column indices range from $0,....,m-1 $. The function we can use to decompose any layer index $x$ in $C^n$ into its row index $i$ and column index $j$ is using this formula: \\

\begin{equation}
i = \text{floor}(n/x)
\end{equation}
\begin{equation}
j = n\% x
\end{equation}

Same formula can be used to decompose the index of any layer into its row and column indices on the condition that the layer's size is a perfect square. The concept can be shown in the figure below:

\begin{figure}[H]
  \includegraphics[width=\linewidth]{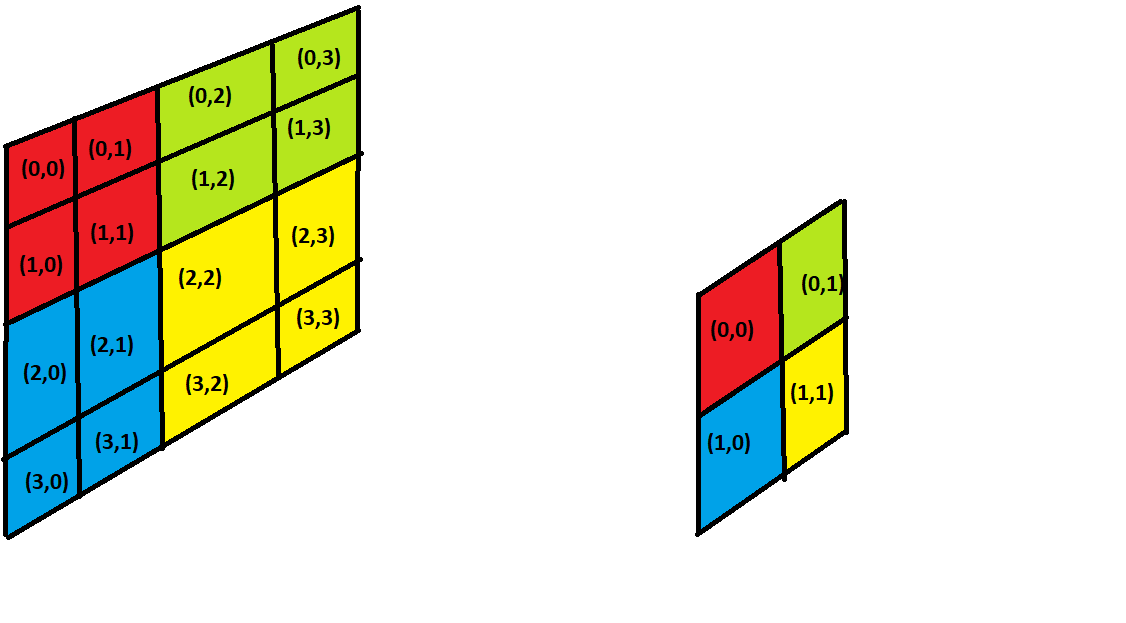}
  \caption{An example of the pooling map from a 4 by 4 layer to a 2 by 2 layer with indices. Same colors means the neurons are connected between the two layers. }
  \label{fig:mapneuron}
\end{figure}

\subsection{Defining the weights}

Now let us define the set of weights as the following map $W : C^n \times C^m \to [0,1]$. Take an element of $C^n$ and call it by its index $x$. $x$ can be decomposed into $(i_1,j_1)$ row and column indices. Similarly, take an element of $C^m$ and call it by its index $y$. $y$ can be decomposed into $(i_2,j_2)$ row and column indices. Define a new value as the ratio $ r = \frac{\sqrt{n}}{\sqrt{m}} $. We will define the map element-wise as follows:

\begin{equation}
w(x,y) = \left\{\begin{array}{lr}
c, & \text{if } \text{floor}(i_1 / r) = i_2, \text{floor}(j_1 / r) = j_2  \\
0, & \text{if } \text{otherwise} \\
\end{array}\right\}
\end{equation}

The map can be generalized to include more overlapping connections between neurons by the following modification:

\begin{equation}
w(x,y) = \left\{\begin{array}{lr}
c, & \text{if } |\text{floor}(i_1 / r) - i_2| \leq v, |\text{floor}(j_1 / r) - j_2| \leq v  \\
0, & \text{if } \text{otherwise} \\
\end{array}\right\}
\end{equation}

Where $c$ is just some constant weight value, and $v$ can be called the connectivity factor and is some integer less than $\sqrt{n} - 1 $, and controls the spread of the connection to each neuron. This can be shown using the following figure: \\

Using this weight connection configuration, we can reduce the dimensions of the input while preserving some form of a distance metric in a process similar to pooling.

\end{document}